\documentclass[letterpaper]{article} 
\usepackage{aaai23}  
\usepackage{times}  
\usepackage{helvet}  
\usepackage{courier}  
\usepackage[hyphens]{url}  
\usepackage{graphicx} 
\usepackage{natbib}  
\usepackage{caption} 
\usepackage{algorithm}

\usepackage{algpseudocode}
\usepackage{makecell}
\usepackage{tikz}
\usepackage{pgfplots, pgfplotstable}

\usepackage{newfloat}
\usepackage{listings}
\DeclareCaptionStyle{ruled}{labelfont=normalfont,labelsep=colon,strut=off} 
\floatstyle{ruled}
\newfloat{listing}{tb}{lst}{}
\floatname{listing}{Listing}
\title{Solving the HP model with Nested Monte Carlo Search}
\author{
      Milo Roucairol,
      Tristan Cazenave
      }

\affiliations {
    Universit\'e   Paris-Dauphine, PSL Research University, CNRS, UMR  7243, LAMSADE, F-75016 Paris, France\\
    milo.roucairol@dauphine.psl.eu,
    tristan.cazenave@dauphine.psl.eu
}

\begin{document}

\maketitle

\begin{abstract}
In this paper we present a new Monte Carlo Search (MCS) algorithm for finding the ground state energy of proteins in the HP-model. We also compare it briefly to other MCS algorithms not usually used on the HP-model and provide an overview of the algorithms used on HP-model. \\
The algorithm presented in this paper does not beat state of the art algorithms, see PERM \cite{hsu_review_2011}, REMC \cite{thachuk_replica_2007} or WLRE \cite{wust_optimized_2012} for better results.

\end{abstract}
\section{Introduction}

Monte Carlo search algorithms have proven to be powerful as game playing agents, with recent successes like AlphaGo\cite{Silver2016MasteringTG}. These algorithms have the advantage of only needing an evaluation function for the final state of the space they explore. 

Protein folding is crucial to our understanding of biology and designing drugs, however, trying our algorithms directly on accurate models could be counterproductive.
In this paper, we use a new MCS algorithm to fold proteins in a simplified lattice based model called the HP model.

First we will present the protein folding and the HP model, then the different algorithms we used to explore the problem space and finally the results of our experiments.

\section{The problem}
\subsection{Protein folding}

With recent developments in ARNm technology, it is now possible to incite cells to produce a specific protein \cite{gros_molecular_1961}, like the spike protein used in COVID-19 vaccines. Unfortunately, deducing the shape a protein will take based given the amino acids sequence is not obvious nor trivial. That is a reason protein folding is a very important problem in molecular biology and medicine.

Proteins are chains of amino acids (primary structure), they can fold in many different ways, the secondary structure is the shape the protein will take at a local level (a coil for example), the tertiary structure is the global shape of the protein with less discernible patterns, finally the quaternary structure is how a protein can assemble with another. Here we are interested in predicting the ground state energy folding (secondary and tertiary structure) from the primary structure. Many forces drive the folding, which prevents the creation of a very accurate simulator, the main driving force is the hydrophobic one.\\

One can not approach protein folding without mentioning DeepMind's AlphaFold \cite{jumper_highly_2021}. Placing first at the Critical Assessment of Techniques for Protein Structure Prediction in 2018 and 2020, it is the best program for protein structure prediction yet.
AlphaFold uses machine learning on a large protein database to train neural networks, in addition to physics based rules in order to predict the folding of a protein.

AlphaFold is the greatest achievement to protein folding prediction in decades, but the research is not over yet. AlphaFold accuracy can still be perfected, and by using neural networks the explainability is low and the model may not be able to predict structure of proteins never seen before.
Our objective here is to provide a better algorithm for Monte Carlo physics simulation, which may be more explainable than AlphaFold, but is currently way less accurate.

\subsection{HP model}

The main idea behind the creation of the HP model is that the Hydrophobic-Polar (HP) force is the main force driving the folding of a protein, thus it is the only one used here.\\

The HP model is a very simplified lattice based model for protein folding, it exists in 2D and 3D versions.

In the HP model, proteins are represented as a chain of H and P residues (amino acids), the chain is then folded onto a grid, two residues can not share the same positions. The residue contacts determine the energy of a chain, usually, the reward for an H-H connection is -1, and 0 for H-P and P-P contacts in a context of minimisation (since the ground energy state is the state with the least potential energy). Other rewards can be used to obtain different results or guide the search.

\subsection{State of the art on HP model}

The HP model was introduced in 1985 by Ken Dill \cite{dill_theory_1985}, it has seen a number of algorithms trying to solve it. All of the best performing algorithms on the HP model are Monte Carlo based, policy learning using neural networks or reinforcement learning like NRPA \cite{Rosin2011} led to poor results. In these Monte Carlo algorithms, we can identify two types, the chain growth algorithms and the replica exchange ones.

The chain growth methods add the residues one after the other, next to the previous one, it is similar to a self avoiding walk and it is the method we used in our own algorithm.

The replica exchange methods use pull moves, pulling the chain at one point by rotating a residue around one of its neighbors, symmetrically rotating a part of the chain or pulling from one side of the chain. This means the entirety of the chain is present on the lattice at any given state of the research, and is in a physically possible conformation, this method is used in simulated annealing like Monte Carlo algorithms. To see a representation of these moves check Chris Thchuk, Alena Shmygelska and Holger H Hoos REMC article \cite{thachuk_replica_2007}.\\

Here is a short review of the methods we encountered.\\

1) PERM :
Initially used on Self Avoiding Walks (SAW), PERM is a chain growth algorithm and was used on the HP model by Peter Grassberger in 1997 \cite{grassberger_pruned-enriched_nodate}. It stands for Pruned Enriched Rosenbluth Method, the idea is to explore the possible chains uniformly with a bias on the immediate gain, cutting (pruning) branches leading to too few choices and poor performances, and cloning (enriching) branches that lead to great results.
PERM has seen many new versions until 2011 \cite{hsu_review_2011}, mainly proposed by its creator, Peter Grassberger. It still is one of the best algorithms available but has been outperformed by pull-moves based more recent algorithms.\\

2) REMC : Introduced in 2007 by Chris Thachuk, Alena Shmygelska and Holger H Hoos \cite{thachuk_replica_2007}, the Replica Exchange Monte Carlo algorithm uses pull moves and simulated annealing. That algorithm keeps only a certain number of replicas (it was determined the best number of replicas for the 3D HP-model was 2), each with a given temperature. At each step the algorithm mutates each replica with a Monte-Carlo Search using the pull moves, the probabilities of keeping a mutation are decided by the score gain (energy loss) of the mutated state and the temperature. Then, once each state is produced through mutation, the replicas are then again swapped probabilistically according to their score (energy) and temperatures.\\

3) Wang-Landau sampling :
Introduced in 2012 on the HP model, the Wang-Landau sampling (WLS) \cite{wust_optimized_2012} method seems to be the new best algorithm for solving the HP model. It is a replica-exchange (simulated annealing) algorithm that uses the same pull moves as the REMC, but also uses moves consisting in cutting and joining of the molecule (thus reallocating all the residues according to their position), together they are named the Monte-Carlo trial moves. With these moves, the WLS explores the conformation space to estimate a histogram of the energies of these conformations. With this histogram, the WLS can then direct the exchange of the replicas.

\section{Algorithm}
\subsection{Biased Growth}
In a similar way to the PERM algorithm \cite{hsu_review_2011}, we try to favor the immediate reward when building/folding the molecule. To do this, we use biased playouts, the chances of selecting a move $m$ from $M$ the possible moves with an immediate gain $G$ follows a softmax distribution with $b$ the bias factor: 

$\frac {exp(G[m]*b)}{sum([exp(G[i]*b) for \; i \; in \; M])}$.\\

$M_{c\mbox -state}$ denotes the legal moves available from the state $current\mbox -state$.\\
$G_{M_{c\mbox -state}}$ denotes the immediate gains of each legal moves available from the state $current\mbox -state$.\\

\begin{algorithm}[tbh]
  \begin{algorithmic}[1]
  \Function{playout}{$c\mbox -state$, $b$}
     \State{$ply \leftarrow 0$}
     \State{$seq \leftarrow \{\}$ }
     \While{$c\mbox -state$ is not terminal}
      \State{$gains \leftarrow G_{M_{c\mbox -state}}$}
      \State{$move \leftarrow softMaxChoice(M_{c\mbox -state}, gains*b)$}
      \State{$current\mbox -state \leftarrow play(c\mbox -state, move)$ }
      \State{$seq[ply] \leftarrow move$}
      \State{$ply += 1$}
     \EndWhile
     \State{}
     \Return{$score(current\mbox -state), seq$}
  \EndFunction
\end{algorithmic}
\caption{\label{playout}The biased growth playout algorithm.}
\end{algorithm}

\subsection{Nested Monte Carlo Search}

\begin{algorithm}[tbh]
\begin{algorithmic}[1]
\Function{NMCS}{$c\mbox -state$, $l$, $b$}
    \If{$l = 0$}
     \Return{$playout(c\mbox -state, b)$}
    \Else
     \State{$best\mbox -score \leftarrow - \infty$}
     \State{$best\mbox -sequence \leftarrow []$}
     \State{$ply \leftarrow 0$}
    \While{$c\mbox -state$ is not terminal}
    \For{{\bf each} $move$ in $M_{c\mbox -state}$}
     \State{$n\mbox -st \leftarrow play (c\mbox -state, move)$}
     \State{$(score, seq) \leftarrow $ \Call{NMCS}{$n\mbox -st, l-1, b$} }
    \If{$score \ge best\mbox -score$}
     \State{$best\mbox -score \leftarrow score$}
     \State{$best\mbox -sequence[ply..] \leftarrow move + seq$}
    \EndIf
      \EndFor
      \State{$next\mbox -move \leftarrow best\mbox -sequence[ply] $}
      \State{$ply \leftarrow ply+1 $}
      \State{$c\mbox -state \leftarrow play(c\mbox -state, next\mbox -move)$}
       \EndWhile
       \State{}
       \Return{($best\mbox -score, best\mbox -sequence$)}
  \EndIf
\EndFunction
\end{algorithmic}
\caption{\label{NMCS}The NMCS algorithm.}
\end{algorithm}

NMCS \cite{CazenaveIJCAI09} is a Monte Carlo Search algorithm that recursively calls lower level NMCS on children states of the current state in order to decide which move to play next, the lowest level of NMCS being a random playout, selecting uniformly the move to execute among the possible moves. A heuristic can be added to the playout move choices, and it is the case here with the biased growth playouts. \\

Algorithm \ref{NMCS} gives the NMCS algorithm, $l$ is the nesting level and $b$ the playout bias.

\subsection{Lazy Nested Monte Carlo Search}

\begin{algorithm}[h]
\begin{algorithmic}[1]

    \State{$tr \leftarrow []$}
  \Function{LNMCS}{$c\mbox -st$, $l$, $b$, $p$, $r$}
    \If{$level = 0$}
     \Return{$\Call{playout}{c\mbox -st, b}$}
    \Else
    \State{$best\mbox -score \leftarrow - \infty$}
     \State{$best\mbox -sq \leftarrow []$}
     \State{$ply \leftarrow 0$}
    \While{$c\mbox -st$ is not terminal}
    \For{{\bf each} $move$ in $M_{c\mbox -state}$}
     \State{$n\mbox -st \leftarrow play (c\mbox -st, move)$}
    
     \For{$i$ in $0..p$}
     \State{$(playoutSc, \_) \leftarrow playout(n\mbox -st, b)$}
     \State{$es \leftarrow es + playoutSc/p$}
     \EndFor
     
     \If{$tr.length() < c\mbox -st.nbplay +1$}
     \State{$tr.push(0.0)$}
     \EndIf
     
     \If{$tr[c\mbox -state.nbplay] < es$}
     \State{$tr[c\mbox -state.nbplay] = es$}
     \EndIf
     
     \If{$es < ratio*tr[c\mbox -st.nbplay]$}
     \State{$(sc, sq) \leftarrow $\Call{LNMCS}{$c.1, 0, b, p, r $}}
     \Else
     \State{$(sc, sq) \leftarrow $\Call{LNMCS}{$c.1, l-1, b, p, r $}}
     \EndIf
     
    \If{$sc \ge best\mbox -score$}
    \State{$best\mbox -score \leftarrow sc$}
     \State{$best\mbox -sq[ply..] \leftarrow move + sq$}
    \EndIf
      \EndFor
              \State{$next\mbox -move \leftarrow best\mbox -sq[ply] $}
      \State{$ply \leftarrow ply+1 $}
      \State{$c\mbox -st \leftarrow play(c\mbox -st, next\mbox -move)$}
       \EndWhile
       \State{}
       \Return{($best\mbox -score, best\mbox -sq$)}
   \EndIf
   \EndFunction
\end{algorithmic}
\caption{\label{LNMCS}The Lazy NMCS algorithm.}
\end{algorithm}

The lazy NMCS inherits its main features from the NMCS, but solves an obstacle encountered on this problem.
Solving the 3D HP model with the NMCS requires using a level of 4 at least, however, it requires computing many 3 level NMCS, already very costly, one for each possible move the level 4 NMCS can make. 
The main idea behind the lazy NMCS is that there are moves that lead to low potential states, to do so, we estimate the potential of a state by launching a number of biased growth playouts and calculating the mean of their scores, then we compare that score to a threshold (relative to the number of moves already done) calculated from the previous estimations to decide if we want to expand the search tree from this state, or prune it.
To update the pruning threshold, it is possible to use a mean, a median or a max from the previous estimations, here we use the max as it gave the best results on these problems.

In the following pseudocode in algorithm \ref{LNMCS}, $p$ is the number of playouts used to evaluate a state and $r$ is the ratio to the threshold a state will be pruned on. $l$ is the nesting level and $b$ is the playout bias.

From line 9 to line 14, the state is evaluated with the mean of $p$ playouts.

From line 13 to line 15, the threshold list is extended on the first entry of a new molecule length, this step is not needed if the list is initialized with the right size from the start for problems we know the maximum number of moves that will be played.

From lines 16 to 18, the threshold is updated with the evaluation.

From lines 19 to 23, it is decided with the evaluation, the pruning ratio and the corresponding threshold if the search will be costly or not.

As you can see there is only one FOR loop iterating over the moves in this implementation of the LNMCS, it means the evaluation is incomplete when the algorithm decides whether to prune the first branches or not. This is a minor flaw in this version of the algorithm and it is easily fixed by ulterior versions (along with other shortcomings). Nonetheless, the experiments were made with this "prototype" version of the algorithm.

\section{Results}
\subsection{Lazy Nested Monte Carlo Search}

We conducted experiments on the 10 molecules with 48 mers from the benchmark we can find in Holger's \cite{thachuk_replica_2007} and Hsu's \cite{hsu_review_2011} work.

\begin{table}[h]
\begin{center}
\begin{tabular}{|c c c|} 
 \hline
 ID & molecule & -E* \\
  \hline
1 & \makecell{HPHHPPHHHHPHHHPPHHPPHPH \\ HHPHPHHPPHHPPPHPPPPPPPPHH} & 32 \\
\hline
2 & \makecell{HHHHPHHPHHHHHPPHPPHHPPH \\ PPPPPPHPPHPPPHPPHHPPHHHPH} & 34 \\
\hline
3 & \makecell{PHPHHPHHHHHHPPHPHPPHPHH \\ PHPHPPPHPPHHPPHHPPHPHPPHP}  & 34 \\
\hline
4 & \makecell{PHPHHPPHPHHHPPHHPHHPPPH \\ HHHHPPHPHHPHPHPPPPHPPHPHP}  & 33 \\
\hline
5 & \makecell{PPHPPPHPHHHHPPHHHHPHHPH \\ HHPPHPHPHPPHPPPPPPHHPHHPH}  & 32 \\
\hline
6 & \makecell{HHHPPPHHPHPHHPHHPHHPHPP \\ PPPPPHPHPPHPPPHPPHHHHHHPH}  & 32 \\
\hline
7 & \makecell{PHPPPPHPHHHPHPHHHHPHHPH \\ HPPPHPHPPPHHHPPHHPPHHPPPH}  & 32 \\
\hline
8 & \makecell{PHHPHHHPHHHHPPHHHPPPPPP \\ HPHHPPHHPHPPPHHPHPHPHHPPP}  & 31 \\
\hline
9 & \makecell{PHPHPPPPHPHPHPPHPHHHHHH \\ PPHHHPHPPHPHHPPHPHHHPPPPH}  & 34 \\
\hline
10 & \makecell{PHHPPPPPPHHPPPHHHPHPPHP \\ HHPPHPPHPPHHPPHHHHHHHPPHH}  & 33 \\
 \hline
\end{tabular}
\caption{\label{BENCHMARK} The benchmarks molecules and their speculated lowest energy state}
\end{center}
\end{table}

To obtain our results, we used the lazy NMCS with a timeout of 150s, if the algorithm has not found a conformation with the lowest known energy before the end of the timeout then we restart the algorithm until that happens.
In our experiments, the playout biased growth gives an immediate gain of 1 to any legal H-H connection, but also a penalty of -0.2 to the HP connections, this was made in order to incite the biased growth to keep a maximum of H mers open to future connections, we did not experiment on that variable.\\ 

Molecule 4 used a lazy NMCS with a threshold based on the mean of the evaluation playouts with the following parameters:\\ 

\begin{center}
\begin{tabular}{|c c|} 
 \hline
 level & 4  \\ 
 \#eval playouts & 10 \\
 pruning ratio & 0.97  \\
 playout bias & 20  \\
 \hline
\end{tabular}
\end{center}

The other molecules used a lazy NMCS with a threshold based on the best average from a batch of evaluation playouts with the following parameters:\\ 

\begin{center}
\begin{tabular}{|c c|} 
 \hline
 level & 5  \\ 
 \#eval playouts & 20 \\
 pruning ratio & 0.9  \\
 playout bias & 20  \\
 \hline
\end{tabular}
\end{center}

Different methods of evaluation and pruning can greatly change the performance of the algorithm, and some methods can be ineffective on a set of molecules while being capable on another set.

\begin{table}[h]
\begin{center}
\begin{tabular}{|c c c|} 
 \hline
 ID & mean time & interquartile \\ [0.5ex] 
 \hline
 1 & 5.5 & 5  \\ 
 2 & 12.5 & 15.5 \\
 3 & 10 & 14  \\
 4 & 20 & 25  \\
 5 & 10 & 10  \\
 6 & +- 180 & -   \\
 7 & +- 60 & -  \\
 8 & 13.5 & 12  \\
 9 & +- 120 & -   \\
 10 & 7 & 9   \\
 \hline
\end{tabular}
\caption{\label{results} Mean time and interquatile (in minutes) of LNMCS on the benchmark molecules}
\end{center}
\end{table}
These results are displayed in minutes and were obtained on a 3.50GHz Intel core i5-6600K CPU.\\

Our LNMCS performed very poorly on molecules 6 and 7, we were not able to gather enough data to compute the statistics. This was unexpected since only molecules 4 and 9 are difficult for PERM (the state of the art chain growth algorithm) to solve according to Grassberger and Hsu's latest paper \cite{hsu_review_2011}, and molecule 4 posed less problems. However, the lazy NMCS could attain the second best energy level very reliably in less than 150s in half of the launches with both molecules.

LNMCS was also able to easily reach the second best level of energy on molecule 9 but could only reach the optimal state every 2 hours approximately, that result was expected since PERM encounters difficulties with that molecule too.

\subsection{Other MCS algorithms}

We tried to solve The HP model with a variety of different Monte Carlo algorithms. We only applied these algorithms to the first molecule from the benchmark (referred as "the molecule" in this section), the results presented in this section are only to give an idea of these algorithms' performances and do not necessarily reflect their potential on the HP model.\\

\subsubsection{Nested Monte Carlo Search}

The good performances of the NMCS \cite{CazenaveIJCAI09} compared to the other algorithms presented in this section is what decided us to try to improve it for this problem into the LNMCS: the NMCS was able to find the optimal value on the molecule in less than 10mn.

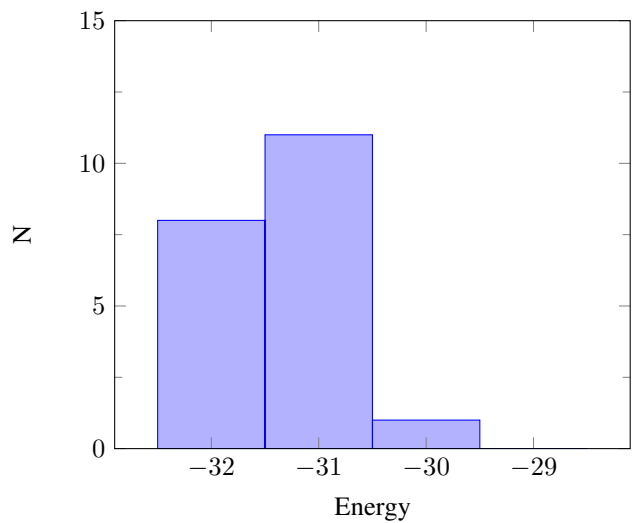
\begin{figure}
\begin{tikzpicture}
\begin{axis}[
    ymin=0, ymax=15,
    minor y tick num = 1,
    area style,
    xlabel={Energy},
    ylabel={N},
    ]
\addplot+[ybar interval,mark=no] plot coordinates { (-28.5, 0) (-29.5, 1) (-30.5, 11) (-31.5, 8) (-32.5, 0) };
\end{axis}

\end{tikzpicture}
\caption{\label{LNMCS_H} Energy distribution with the  Lazy NMCS}
\end{figure}

\begin{figure}
\begin{tikzpicture}
\begin{axis}[
    ymin=0, ymax=15,
    minor y tick num = 1,
    area style,
    xlabel={Energy},
    ylabel={N},
    ]
\addplot+[ybar interval,mark=no] plot coordinates { (-28.5, 0) (-29.5, 3) (-30.5, 14) (-31.5, 3) (-32.5, 0) };
\end{axis}
\end{tikzpicture}
\caption{\label{NMCS_H} Energy distribution with the NMCS}
\end{figure}
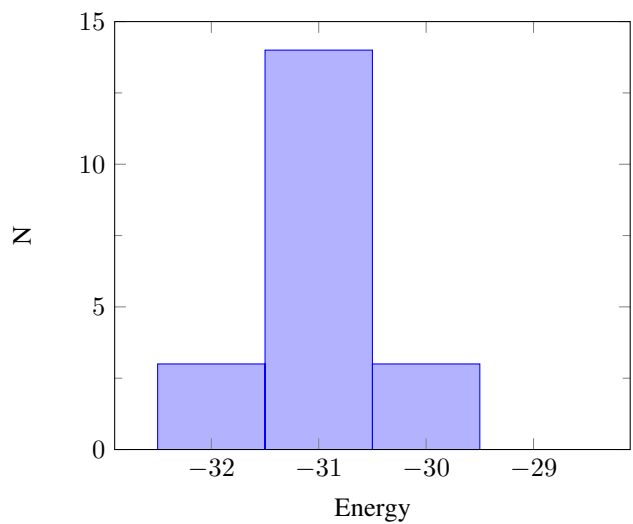

In Figure \ref{NMCS_H} and Figure \ref{LNMCS_H} we compare performances of the level 5 NMCS and the level 5 LNMCS with a ratio of 0.9 on molecule 1, with both a playout bias of 20 over 20 runs with a 150s timeout.\\

As you can see on these figures, the LNMCS provides a substantial performance gain over the NMCS on this problem. Lowest energy conformations are way sparser than the second lowest energy conformations (about an order of magnitude or two), being able to reach them 4 times as often is a great improvement.

\subsubsection{Nested Rollout Policy Adaptation}

The NRPA \cite{Rosin2011} is similar to the NMCS, the main difference is that the NRPA learns a policy to decide which move play during playouts, that policy discovery is interesting for many problems which are too complex to implement a man made policy (like we did here). On many problems the NRPA outperforms the NMCS, however in our case it was not able to do. The performance of the NRPA and GNRPA \cite{Cazenave2020GNRPA} are widely dependent on the move representations, here it is the number corresponding to the number of residues already placed and its direction and NRPA and GNRPA with a bias of 20 were not able to reach the optimal energies (-28 or -29 for the GNRPA when -32 is the best known). Other moves representations were tried, using the last few previous moves instead of the number of residues already placed for example, but none were able to provide better performances. Our inability to obtain good results with NRPA does not mean it is impossible to solve this problem with it.\\

\subsubsection{Greedy Best First Search with playouts}

The Greedy BFS \cite{doran1966experiments} is a simple search algorithm that uses a ranked list of the nodes to open according to their scores given by an evaluation function.
Iteratively, the Greedy BFS opens the best node from the list and launches the evaluation function on every child of this node to insert them in the ranked list. Here we evaluate the children with their results with one or multiple playouts.
This method converges rapidly to a "good enough" solution (a local minimum), -29 when the best known is -32 on molecule 1, but then improves very little, it is due to a large number of good scoring states that do not lead to an optimal solution, making the search too exhaustive. Pruning the search tree could improve the results of this method.\\

\subsubsection{Upper Confidence bounds applied to Trees}

 UCT \cite{hutchison_bandit_2006} is used on game playing and usually provides good results. It iteratively starts from the initial state and launches playouts, based on the results of each playout, the value used to determine which moves should be taken is updated. UCT does not work well on the HP-model without biased growth (achieving about the same scores as a single biased playout, -18), with biased growth it achieves scores around -28/-29 on molecule 1, like the other non NMCS algorithms discussed here.

\section{Conclusion}

While the Lazy NMCS algorithm likely does not outperform the state of the art algorithms, it has the advantage to be easier to implement and may be applicable to more problems. It is also shown to be an improvement on the NMCS algorithm for this specific problem. In future works, we aim to apply it to other problems and find ways to improve its performance. 

\bibliography{aaai23}

\end{document}